\documentclass[letterpaper, 10 pt, conference]{ieeeconf}
\IEEEoverridecommandlockouts
\usepackage{amsmath,amsfonts,amssymb}
\usepackage{algpseudocode}
\usepackage{algorithm}
\usepackage{array}
\usepackage{svg}
\usepackage{textcomp}
\usepackage{stfloats}
\usepackage{url}
\usepackage{verbatim}
\usepackage{graphicx}
\usepackage{cite}
\usepackage{wrapfig}
\usepackage{multirow}
\usepackage{microtype}
\usepackage{tabularx,ragged2e}
\usepackage{booktabs}
\usepackage[letterpaper, left=0.667in, right=0.667in, bottom=0.597in, top=0.792in]{geometry}
\usepackage{fancyhdr}
\usepackage{threeparttable}
\usepackage{siunitx}

\makeatletter
\let\NAT@parse\undefined
\makeatother
\usepackage{hyperref}

\begin{document}

\title{\LARGE \bf 
Learning Semantic Traversability with Egocentric Video \\and Automated Annotation Strategy
}

\author{Yunho~Kim$^{*, 1, 2,}$\textsuperscript{\textdagger},
        Jeong Hyun~Lee$^{*, 1}$,
        Choongin~Lee$^1$,
        Juhyeok~Mun$^1$,\\
        Donghoon Youm$^1$,
        Jeongsoo~Park$^1$,
        and~Jemin~Hwangbo$^{**, 1}$
\thanks{This work was supported by Samsung Research Funding \& Incubation Center for Future Technology at Samsung Electronics under Project Number SRFC-IT2002-02.}
\thanks{$*$ equal contribution, $**$ corresponding author}
\thanks{$1$ Robotics and Artificial Intelligence Lab, KAIST, Daejeon, South Korea}
\thanks{$2$ Neuromeka, Seoul, South Korea}
\thanks{\textsuperscript{\textdagger} Substantial part of the work was carried out during his stay at $1$}
\thanks{\tt\footnotesize yunho.kim@neuromeka.com, joshualee@kaist.ac.kr, jhwangbo@kaist.ac.kr}
}

\maketitle
\thispagestyle{empty}
\pagestyle{empty}

\thispagestyle{fancy}
\fancyhf{}
\renewcommand{\headrulewidth}{0pt}

\begin{abstract}
For reliable autonomous robot navigation in urban settings, the robot must have the ability to identify semantically traversable terrains in the image based on the semantic understanding of the scene. This reasoning ability is based on semantic traversability, which is frequently achieved using semantic segmentation models fine-tuned on the testing domain. This fine-tuning process often involves manual data collection with the target robot and annotation by human labelers which is prohibitively expensive and unscalable. In this work, we present an effective methodology for training a semantic traversability estimator using egocentric videos and an automated annotation process. Egocentric videos are collected from a camera mounted on a pedestrian's chest. The dataset for training the semantic traversability estimator is then automatically generated by extracting semantically traversable regions in each video frame using a recent foundation model in image segmentation and its prompting technique. Extensive experiments with videos taken across several countries and cities, covering diverse urban scenarios, demonstrate the high scalability and generalizability of the proposed annotation method. Furthermore, performance analysis and real-world deployment for autonomous robot navigation showcase that the trained semantic traversability estimator is highly accurate, able to handle diverse camera viewpoints, computationally light, and real-world applicable. The summary video is available at \url{https://youtu.be/EUVoH-wA-lA}.
\end{abstract}


\section{Introduction}
Mobile robots should possess autonomous navigation capabilities as it is an essential feature for various applications, including package delivery, factory inspection, urban security, and search and rescue. In the realm of autonomous navigation, various problem domains exist. Unlike the majority of prior research, which primarily addresses autonomous off-road navigation in field environments\cite{maturana2018offRoadNav, wellhausen2019point, kahn2021badgr, frey2023fastTrav, lee2023cropLSTM} or autonomous on-road navigation in urban settings\cite{cordts2016cityscapes, barnes2017autoDriveSeg}, our focus lies in tackling the challenges of autonomous off-road navigation within urban environments. 

\begin{figure}[!t]
\centering
\includegraphics[width=\linewidth]{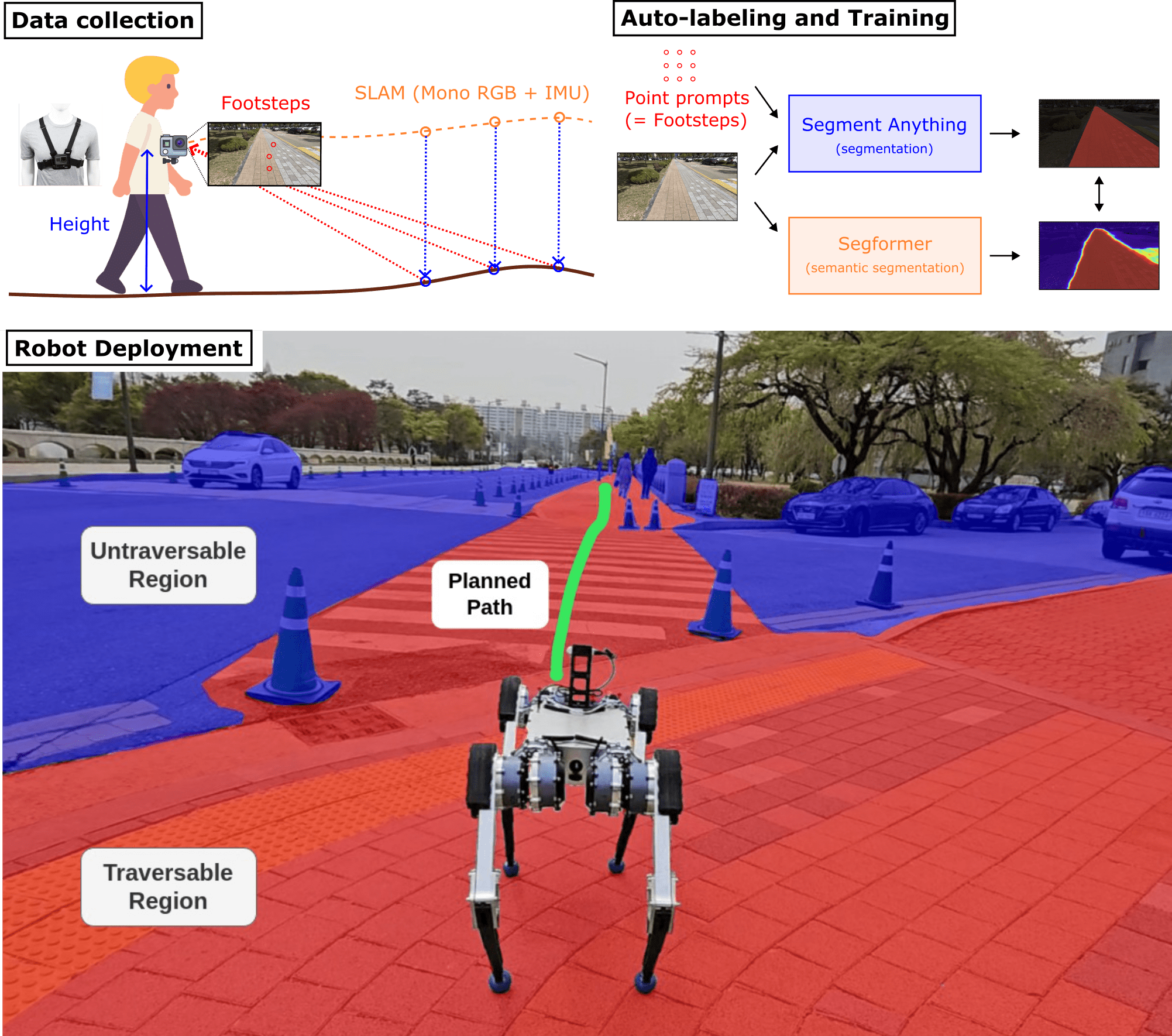}
\caption{Overall framework}
\vspace{-0.5cm}
\label{fig:overall framework}
\end{figure}

Robot navigation frameworks are often built with several modules, one of which is the path planner. The path planner finds a path on the map\cite{erni2023multiModalMap} by considering the geometric and semantic traversability of each point. 
Geometric traversability is computed using either analytic techniques\cite{wermelinger2016analyticFilter} or learning-based methods\cite{kim2022leggedLocalPlanner, frey2022volumTrav} from a map constructed with depth cameras or lidar, which contains geometric information such as grid occupancy and terrain height. Semantic traversability is often evaluated using semantic segmentation predictions from semantic segmentation models\cite{xie2021segformer, cheng2022mask2former} trained on large-scale datasets containing RGB images and semantic class annotations. Engineers assign a heuristic objective value to each terrain class based on their preference\cite{maturana2018offRoadNav, guan2022gaNav, roth2023viplanner}. For example, if the engineer wants the robot to have a preference $Sidewalk > Grass > Road$, the objective values for each terrain class can be set as $Sidewalk=1,\, Grass=0.5,\, Road=0.25$, indicating high values for high preference.

However, directly computing semantic traversability based on semantic segmentation predictions limits its performance to that of the trained semantic segmentation model. Although the deployed semantic segmentation models are trained on large-scale datasets, they often show limited generalization performance due to a significant domain gap between the training and testing datasets\cite{sankaranarayanan2018semsegDA}. Specifically, the models are vulnerable to out-of-distribution (OOD) RGB images (e.g., a model trained on autonomous-driving datasets results in a strong bias toward the camera viewpoint of street vehicles and shows limited performance in non-vehicle regions such as pedestrian areas) and OOD terrain classes (e.g., a model trained to recognize $Grass,\, Sand,\, and\,\, Rock$ cannot be directly used to predict classes such as $Sidewalk,\, Road,\, and\,\, Crosswalk$). 

Because incorrect classifications of terrain classes can deteriorate navigation behavior, the trained semantic segmentation models are usually fine-tuned before being deployed with manually collected and annotated data from the testing domain\cite{roth2023viplanner}. However, the process of manually preparing the semantic segmentation dataset is highly expensive, labor-intensive, and not scalable. For rapid development of the semantic traversability estimator, a more scalable data collection method and an automated labeling technique are necessary. Furthermore, the model design of the semantic traversability estimator should be adaptable enough to handle OOD terrain classes and OOD visual features.

To this end, we present an effective methodology for training a neural network that predicts semantic traversability using a highly scalable data acquisition method and an automated annotation strategy without the need for manual labeling (Figure \ref{fig:overall framework}). Instead of using the target robot to gather data, we utilize egocentric videos obtained by mounting a camera on a person's chest and capturing the egocentric perspective of a pedestrian crossing an urban area. The collected egocentric videos are then automatically labeled by prompting a segmentation model\cite{kirillov2023segany} with approximate footsteps extracted using the Monocular Visual-Inertial SLAM\cite{campos2021orb3} and refining the outputs with a large semantic segmentation model\cite{cheng2022mask2former}. The semantic traversability estimator is then obtained by fine-tuning a lightweight semantic segmentation model\cite{xie2021segformer} with the auto-labeled dataset and minimal model architecture modification. We have conducted extensive quantitative and qualitative evaluations to demonstrate the effectiveness of our method in terms of data collection, data annotation, and model training. Specifically, we show that leveraging egocentric video with the proposed automated annotation strategy possesses great potential for highly scalable and worldwide training data collection due to its minimal hardware requirements. Additionally, we demonstrate that fine-tuning the pre-trained semantic segmentation model yields a semantic traversability estimator that is highly accurate, able to handle diverse camera viewpoints, and lightweight (71Hz on a desktop GPU and 16Hz on an onboard embedded GPU). The trained semantic traversability estimator is further deployed for autonomous quadruped robot navigation in an urban environment, showcasing the method's capability for real-world applications.

\section{Related work}
There have been several prior research on learning a semantic traversability estimator (typically parameterized with neural networks) in a self-supervised manner without the need for manual labels. Wellhausen et al.\cite{wellhausen2019point} used the future robot footsteps projected on the current image as traversable point labels for training the traversability estimator. Each point label included ground reaction scores from force-torque sensors as well as corresponding terrain classes. Schmid et al.\cite{schmid2022reconstruct} approached the problem as an anomaly detection problem, training an autoencoder to reconstruct only the traversed area of the image. The traversable region on the image was then identified as an area with a low reconstruction error. Frey and Mattamala et al.\cite{frey2023fastTrav} integrated weak image segmentation masks with automatically derived traversability values from velocity command tracking errors, allowing the traversability estimator to consider semantic cues. Jung et al.\cite{jung2023vstrong} used Segment Anything model (i.e., SAM)\cite{kirillov2023segany} and contrastive learning in self-supervised traversability learning for field environment navigation. 

Our work is closely related to several previous approaches but greatly distinguishes itself in two points. First, we leverage a two-step annotation method in which the labels are first produced from the point-prompt-based segmentation model (i.e., SAM) and then refined with the large semantic segmentation model (i.e., Mask2former). The process results in large traversable \textit{area} labels instead of local \textit{point}\cite{wellhausen2019point} or \textit{segment}\cite{frey2023fastTrav} labels and enables larger supervision signal during training. Furthermore, the additional refinement step allows for more fine-grained and reliable traversability labels by incorporating semantic information and filtering out SAM's overestimated predictions\cite{jung2023vstrong}, which are frequently encountered due to a lack of visual cues. Our extensive evaluations under a variety of environmental conditions demonstrate the effectiveness of the proposed two-step annotation method, as well as its applicability for use in not only field and park-like environments but also urban scenarios. Second, we utilize egocentric videos captured by pedestrians as the data source and demonstrate its capability for the usage of training mobile robots’ traversability estimators. The suggested data acquisition method enables effective and scalable data collection compared to prior research \cite{wellhausen2019point, frey2023fastTrav, jung2023vstrong, schmid2022reconstruct}, which collected data directly by deploying the target robot. To the best of our knowledge, our work is the first to demonstrate that high-performing traversability estimators can be trained with egocentric videos recorded by pedestrians and further deployed for real-robot navigation in urban scenarios.

\section{Method}
In this section, we first describe the data collection and automatic annotation pipeline for creating a dataset to train the semantic traversability estimator, followed by an elaboration on its training with the created dataset. We then illustrate how we integrate the trained semantic traversability estimator into the hierarchical navigation framework and use it for real-world robot autonomy in urban settings.

\subsection{Data collection}
We collect egocentric videos obtained by mounting a GoPro camera on a person's (i.e., volunteer) chest and capturing the egocentric perspective of the volunteer walking around an urban environment. This approach enables effective and scalable data collection compared to previous works\cite{wellhausen2019point, frey2023fastTrav, schmid2022reconstruct, karnan2023sterling, sikand2022visualPF} that rely on data obtained directly from the target robot. Leveraging the target robot itself often requires well-calibrated sensor settings and limits the data acquisition capability based on several constraints such as the number of available robots, the robot's speed, and physically reachable area. The volunteer is instructed to walk around inside the designated area while taking into account their visual path preference, which is consistent with urban regulations such as pedestrians walking on the sidewalk rather than the road and using crosswalks when crossing the road. The volunteer is not restricted from going to specific terrains, such as steep bumps, because our target robot for deploying the semantic traversability estimator is a legged robot that can traverse most regions that humans can traverse\cite{kim2023leggedConstrainedRL}. However, volunteers may be given instructions in advance to avoid certain terrains, taking into account the type of target robot and its controller capability, which prevents the corresponding terrain from being labeled as traversable. For example, if the target robot is a wheeled robot, the volunteer may be told to avoid stairs and vertical slopes because the robot cannot overcome them.

\subsection{Automatic annotation}

We develop an automatic annotation pipeline that extracts visually preferred regions to walk in each frame of the collected egocentric videos. First, ORB-SLAM3\cite{campos2021orb3} is used to estimate the camera trajectory—a set of coordinates representing the camera's position and orientation with respect to a fixed global reference frame—from each video clip. We can obtain global camera poses on an absolute scale by running the Monocular Visual-Inertial SLAM with the synchronized IMU sensor built inside the GoPro camera. Because there is not much of a visual difference between every two consecutive frames (in our case, the camera frame rate was 24Hz), we only use camera frames that are identified as keyframes. Let's denote the estimated SE(3) pose trajectory of the camera in a single video clip as ${}^{W}_{}\textbf{T}_{0:N} = [{}^{W}_{}\textbf{R}_{0:N} | {}^{W}_{}\textbf{P}_{0:N}]$, where ${}^{W}_{}\textbf{R}$ and ${}^{W}_{}\textbf{P}$ are each rotation matrix and position vector, $N$ is the number of detected keyframes, and $W$ is the fixed global reference frame that corresponds to the gravity-aligned frame at the initial camera pose. We then obtain the approximate terrain points that the volunteer traversed, utilizing the fact that the camera is mounted on the volunteer's chest as 
\begin{equation}\label{eq:approximate_footstep}
    {}^{W}_{}\textbf{P}_{0:N}^{f} = {}^{W}_{}\textbf{P}_{0:N} - [0, 0, H],
\end{equation}
where $H$ is the measured distance from the ground to the mounted camera when the volunteer stands still. We will refer to these approximate terrain points that the volunteer traversed as \textit{footsteps}. The footsteps are projected into the image space as
\begin{equation}
    \textbf{p}_{i:i+n_i}^{f} = \textbf{K} \cdot \textbf{T}_{CW}^{i} \cdot {}^{W}_{}\textbf{P}_{i:i+n_i}^{f},
\end{equation}
where $C$ is the camera frame, $\textbf{K}$ is the intrinsic camera calibration matrix, $\textbf{T}_{CW}^{i}={}^{W}_{}\textbf{T}_{i}^{-1}$ is the extrinsic transformation matrix from the fixed global reference frame into the $i$th camera frame, and $n_i$ is the number of considered keyframes at the $i$th camera frame. The number of future keyframes $n_i$ to be taken into account for each frame is determined by the predefined time horizon $T'$. In particular, keyframes between $t$ and $t+T'$ time stamps are taken into consideration for the $i$th camera frame that corresponds to the $t$ time stamp. We ignore potential occlusions of footsteps caused by vertical geometry and dynamic obstacles, and we also note that a short time horizon $T'$ (in our case, $T'=3s$) resulted in almost no occlusion. In the future, occluded footsteps can be removed using a 3D reconstruction of the environment.

The semantically traversable area in the current frame that the volunteer intends to traverse is obtained from the projected footsteps. Our key insight is that the semantically preferred region in the image will include the projected footsteps and differ from the semantically undesirable region by visual cues like edges, colors, and textures. These visual cues are important for humans to distinguish between geometrically similar but semantically distinct paths (e.g., a road and a sidewalk, a bicycle track and a pedestrian track), and urban planning regulations also take them into account. To this end, we leverage SAM with its powerful prompting technique to extract the semantically desirable area. SAM, trained on an extremely large-scale dataset of 1 billion images, demonstrates high generalizability across diverse images and robust performance in segmenting regions that align well with given prompts. For each frame, we obtain mask predictions from SAM by passing the model with the corresponding image and projected footsteps as positive point prompts. Although SAM can handle both RGB and gray-scale images, we used gray-scale images since SAM's results were highly color-sensitive. When projecting points from 3D space to image space, their distribution on the image becomes non-uniform, concentrating more densely in regions far from the observer. These non-uniform point prompts may degrade the predicted mask from SAM, and thus $n_p$ (in our case, $n_p=3$) points are sampled from the projected footsteps $\textbf{p}_{i:i+n_i}^{f}$ via farthest point sampling and used for positive point prompts. Additionally, we post-process the raw mask predictions from SAM with an area and contour filter. The area filter selects a mask from the proposals if its area exceeds a certain threshold. The contour filter extracts contours from mask predictions and leaves only the contour with the biggest area. This allows for the removal of tiny, independently segmented pieces.

\begin{figure}[t!]
\centering
\includegraphics[width=0.9\linewidth]{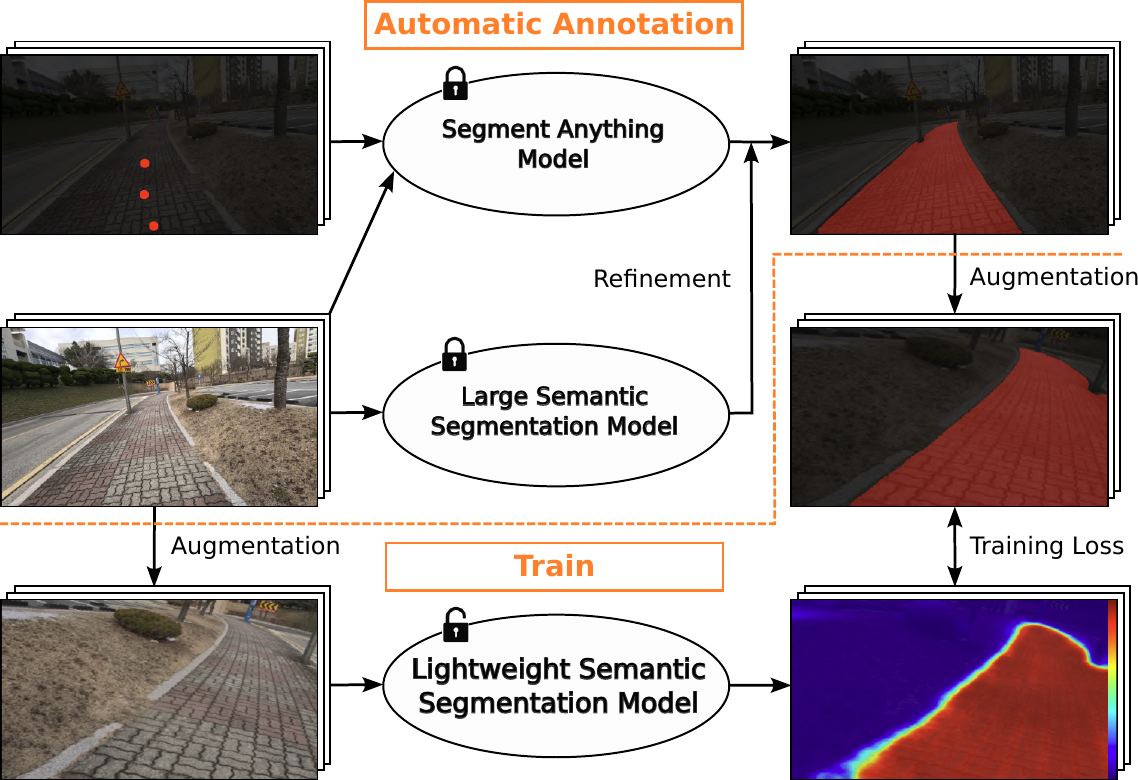}
\caption{Automatic annotation process}
\vspace{-0.5cm}
\label{fig:automatic annotation process}
\end{figure}

We use a large semantic segmentation model \cite{cheng2022mask2former} in conjunction with SAM to produce the semantic traversability labels for the training dataset. This process introduces more detailed semantic information required for urban environment deployment (e.g., \textit{road} is less preferred compared to other traversable terrains) and filters out SAM’s overestimated
predictions. Even though a large semantic segmentation model's computational load makes it unsuitable for high-frequency real-time usage (i.e., Mask2former\cite{cheng2022mask2former} model that we used runs at 1.66Hz on the onboard embedded GPU), this is less important for offline data annotation. From the predictions of Mask2former, segments labeled as \textit{crosswalk} and \textit{road} are added to and removed from the SAM mask predictions, respectively. If projected footsteps are sufficiently included in the remaining masks, the remaining segmented areas are annotated with a value of 1. If not, this is typically a scenario when the volunteer walked on the \textit{road} because it was the best navigation behavior in their current field of view. In this instance, the areas where the SAM mask intersects with areas identified as \textit{road} are given a value of 0.25, whereas the remaining SAM mask areas and areas identified as \textit{crosswalk} are given a value of 1. The annotation process is summarized in Figure \ref{fig:automatic annotation process}.

\subsection{Model training}
We use the dataset acquired through our automatic annotation pipeline to train a semantic traversability estimator. The dataset is composed of tuples of RGB images and pixel-wise traversability values ranging from 0 to 1, with higher values indicating a stronger preference. We leverage neural networks to parameterize the semantic traversability estimator. However, rather than training it from scratch, we model it with a lightweight semantic segmentation model trained on a large-scale dataset (i.e., Segformer\cite{xie2021segformer}) and fine-tune it with the collected dataset. A single 1x1 convolution and sigmoid activation layer is added to Segformer's final layer to reduce the number of channels from the number of segmentation classes to one and restrict the output from 0 to 1. Fine-tuning a pre-trained semantic segmentation model can be thought of as auto-tuning class-wise traversability values and adapting OOD representations caused by visual effects and terrain variations. Previously, these processes, such as class-wise traversability value tuning and semantic segmentation data labeling, were carried out manually\cite{maturana2018offRoadNav, guan2022gaNav, roth2023viplanner}.

The entire model parameters are fine-tuned via gradients derived from L2 loss for non-zero traversability labels and L1 loss with a weighting factor of 0.05 for zero traversability labels. Zero traversability labels are less penalized than non-zero traversability labels because, in our annotation pipeline, zero traversability labels may sometimes include both non-traversable and traversable but not yet traversed areas. The corresponding consideration in the loss function, as well as the model's weak inductive bias from being initialized with the pre-trained semantic segmentation model, allow for robust traversability estimator learning in the presence of minor data noise. Random image augmentations (e.g., flipping, rotation, resize-cropping, and color jittering) are performed during training to deploy the semantic traversability estimator trained on egocentric videos to a variety of camera viewpoints. 

\subsection{Usage for real-world robot navigation}
We deploy the trained semantic traversability estimator for autonomous quadruped robot navigation in an urban environment. Pixel-aligned RGB-D images, obtained from a single RGB-D camera (i.e., Intel RealSense D435) mounted on top of the target robot \textit{Raibo 2}, are used for traversability estimation. The RGB image is first passed to the trained semantic traversability estimator to predict per-pixel semantic traversability values. The per-pixel semantic traversability values and depth values are then converted from the image space to the 3D space and accumulated in 2.5D grid maps (size: 10\si{m} x 10\si{m}, resolution: 0.025\si{m}). We instantiate two separate grip map layers, each for storing grid-cell-wise geometric traversability and semantic traversability. Geometric traversability is computed using terrain heights readily obtainable from per-pixel depth values. We use RRT* to find a path that reaches the target waypoint from the current location. The target waypoints are provided 4m ahead of the robot with a given yaw angle, and the current location is known from visual-inertial SLAM\cite{campos2021orb3}. When determining the path, path length, geometric traversability, and semantic traversability are all considered as part of an objective function.

\section{Experimental Results}
The proposed method for semantic traversability estimation was thoroughly evaluated, both qualitatively and quantitatively.

\subsection{Evaluation of data collection and automatic annotation}
We collected a wide range of egocentric videos and tested whether our automatic annotation method can identify areas that are well-aligned with humans' visual preferences while also demonstrating robust performance across visually diverse video scenarios. The videos were taken in several countries (i.e., South Korea, Japan, and Italy) and cities (i.e., Daejeon, Seoul, Tokyo, Rome, and Venice), and included a variety of semantically traversable regions found in both urban and field environments, such as sidewalks with varying patterns, pedestrian areas, crosswalks, stairs, grassy terrains, snow regions, dirt areas, and others that are visually appealing to traverse but difficult to categorize. This diverse data collection could be done effectively due to the minimal hardware requirement of a single camera rather than the target robot itself with well-calibrated sensor settings. The efficacy of egocentric videos further highlights the potential for scalable dataset construction for semantic traversability estimation.

\begin{figure*}[t!]
\centering
\scalebox{1.0}[0.8]{\includegraphics[width=\linewidth]{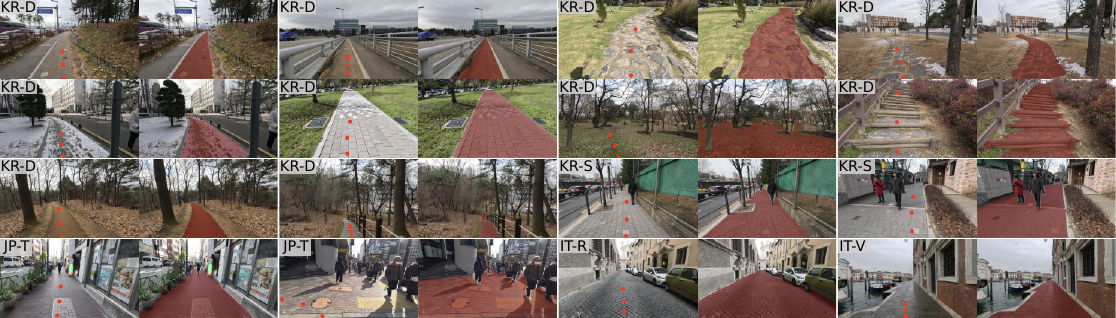}}
\caption{Automatic annotation results (KR-D: Daejeon, South Korea / KR-S: Seoul, South Korea / JP-T: Tokyo, Japan / IT-R: Rome, Italy / IT-V: Venice, Italy)}
\vspace{-0.25cm}
\label{fig:annotation results}
\end{figure*}

\begin{figure}[t!]
\centering
\scalebox{1.0}[0.8]{\includegraphics[width=\linewidth]{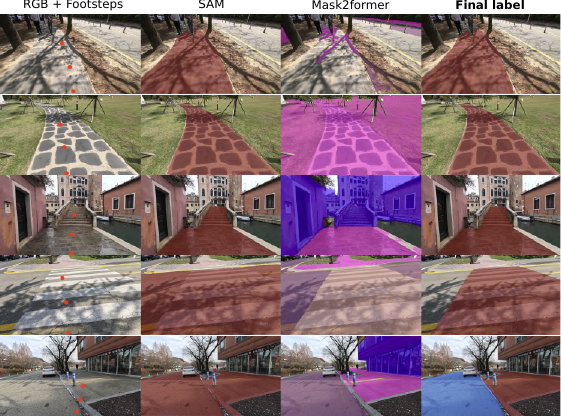}}
\caption{Stepwise annotation results. In Mask2former predictions, only sidewalks, crosswalks, roads, and buildings are color-segmented with pink, brown, purple, and blue if detected. In the final labels, sky-blue and red each indicates a value of 0.25 and 1.}
\vspace{-0.5cm}
\label{fig:stepwise annotation results}
\end{figure}

The annotated results are shown in Figure \ref{fig:annotation results}. Clear traversable area labels consistent with humans' semantic traversability were obtained by expanding the volunteer's future footsteps using SAM. These expanded area labels, as opposed to point labels as in the previous work\cite{wellhausen2019point}, provide rich supervision signals when training the semantic traversability estimator, and will be further explained in Section \ref{subsec:eval_model}. Owing to the SAM's highly generalizable segmentation capability, our automated annotation pipeline showed robust performance in a variety of video scenarios in both field and urban settings. The use of a large semantic segmentation model (i.e., Mask2former\cite{cheng2022mask2former}) in conjunction with SAM resulted in improved final labels that complemented each other (Figure \ref{fig:stepwise annotation results}). SAM is particularly good at segmenting areas that include all footsteps based on visual cues, but it has limited performance in segmenting crosswalks where footsteps are placed not only on the white patterns but also on the road, resulting in overly large area labels. Mask2former, on the other hand, can reliably predict category-level semantic information, particularly for dominant categories such as "road," but not for non-dominant categories. Through the combination of these two models, we were able to obtain final annotations for images where either solely SAM cannot generate (e.g., crosswalk) or Mask2former cannot accurately predict (e.g., stairs, sidewalk, and pedestrian areas), and further achieve labels that include semantic category-level preference (i.e., road is less preferred than other traversed areas). 

\subsection{Evaluation of the trained semantic traversability estimator}\label{subsec:eval_model}
We obtained the semantic traversability estimator by fine-tuning a lightweight semantic segmentation model (i.e., Segformer\cite{xie2021segformer}) with approximately 2 hours and 20 minutes of egocentric videos. Videos were taken inside the KAIST campus (site size approximately 1.1M \si{m^2}, 16.3 times the size of a soccer field), which is the testing region for our robot to perform autonomous navigation. The proposed automatic annotation pipeline produced 57K data tuples from the collected videos, each tuple containing an RGB image and pixel-wise traversability values, which were then randomly divided into train and validation sets with ratios of 0.9 and 0.1. The model was trained for 100 epochs with a batch size of 48 and a learning rate of 1e-5.

\begin{table*}[ht!]
\caption{Quantitative evaluation results on the urban dataset}
\centering
\resizebox{\textwidth}{!}{\begin{tabular}{|l|c|c|cccc|cccc|}
\hline
\multirow{2}{*}{} &
  \multirow{2}{*}{\begin{tabular}[c]{@{}c@{}}Computation \\ time {[}\si{ms}{]}\end{tabular}} &
  \multirow{2}{*}{\begin{tabular}[c]{@{}c@{}}\# of \\ parameters\end{tabular}} &
  \multicolumn{4}{c|}{Same viewpoint} &
  \multicolumn{4}{c|}{Different viewpoint} \\ \cline{4-11} 
 &
   &
   &
  \multicolumn{1}{c|}{Precision} &
  \multicolumn{1}{c|}{Recall} &
  \multicolumn{1}{c|}{IoU} &
  RMSE &
  \multicolumn{1}{c|}{Precision} &
  \multicolumn{1}{c|}{Recall} &
  \multicolumn{1}{c|}{IoU} &
  RMSE \\ \hline\hline
\textbf{{[}Segformer + 1x1 conv{]} + Area labels (Ours)} &
  \textbf{14 / 61} &
  3.7M &
  \multicolumn{1}{c|}{\begin{tabular}[c]{@{}c@{}}\textbf{0.951}\\ (0.927 / 0.975)\end{tabular}} &
  \multicolumn{1}{c|}{\begin{tabular}[c]{@{}c@{}}\textbf{0.956}\\ (0.972 / 0.940)\end{tabular}} &
  \multicolumn{1}{c|}{\textbf{0.903}} &
  \textbf{0.178} &
  \multicolumn{1}{c|}{\begin{tabular}[c]{@{}c@{}}\textbf{0.965}\\ (0.953 / 0.977)\end{tabular}} &
  \multicolumn{1}{c|}{\begin{tabular}[c]{@{}c@{}}\textbf{0.963}\\ (0.975 / 0.950)\end{tabular}} &
  \multicolumn{1}{c|}{\textbf{0.930}} &
  \textbf{0.161} \\ \hline
{[}Segformer + 1x1 conv{]} + Point labels \cite{wellhausen2019point} &
  14 / 61 &
  3.7M &
  \multicolumn{1}{c|}{\begin{tabular}[c]{@{}c@{}}0.818\\ (0.985 / 0.650)\end{tabular}} &
  \multicolumn{1}{c|}{\begin{tabular}[c]{@{}c@{}}0.629\\ (0.260 / 0.997)\end{tabular}} &
  \multicolumn{1}{c|}{0.257} &
  0.540 &
  \multicolumn{1}{c|}{\begin{tabular}[c]{@{}c@{}}0.787\\ (0.989 / 0.584)\end{tabular}} &
  \multicolumn{1}{c|}{\begin{tabular}[c]{@{}c@{}}0.609\\ (0.220 / 0.997)\end{tabular}} &
  \multicolumn{1}{c|}{0.219} &
  0.597 \\ \hline
Autoencoder \cite{schmid2022reconstruct} &
  28 / 97 &
  \textbf{0.6M} &
  \multicolumn{1}{c|}{\begin{tabular}[c]{@{}c@{}}0.610\\ (0.534 / 0.685)\end{tabular}} &
  \multicolumn{1}{c|}{\begin{tabular}[c]{@{}c@{}}0.620\\ (0.529 / 0.711)\end{tabular}} &
  \multicolumn{1}{c|}{0.358} &
  0.597 &
  \multicolumn{1}{c|}{\begin{tabular}[c]{@{}c@{}}0.622\\ (0.579 / 0.666)\end{tabular}} &
  \multicolumn{1}{c|}{\begin{tabular}[c]{@{}c@{}}0.628\\ (0.632 / 0.624)\end{tabular}} &
  \multicolumn{1}{c|}{0.437} &
  0.598 \\ \hline
Segformer + Heuristic \cite{maturana2018offRoadNav, guan2022gaNav, roth2023viplanner} &
  94 / 886 &
  3.7M &
  \multicolumn{1}{c|}{\begin{tabular}[c]{@{}c@{}}0.850\\ (0.750 / 0.950)\end{tabular}} &
  \multicolumn{1}{c|}{\begin{tabular}[c]{@{}c@{}}0.829\\ (0.918 / 0.741)\end{tabular}} &
  \multicolumn{1}{c|}{0.700} &
  0.371 &
  \multicolumn{1}{c|}{\begin{tabular}[c]{@{}c@{}}0.835\\ (0.752 / 0.918)\end{tabular}} &
  \multicolumn{1}{c|}{\begin{tabular}[c]{@{}c@{}}0.810\\ (0.889 / 0.730)\end{tabular}} &
  \multicolumn{1}{c|}{0.696} &
  0.392 \\ \hline
Mask2former + Heuristic \cite{maturana2018offRoadNav, guan2022gaNav, roth2023viplanner} &
  89 / 692 &
  43.9M &
  \multicolumn{1}{c|}{\begin{tabular}[c]{@{}c@{}}0.911\\ (0.861 / 0.961)\end{tabular}} &
  \multicolumn{1}{c|}{\begin{tabular}[c]{@{}c@{}}0.917\\ (0.963 / 0.871)\end{tabular}} &
  \multicolumn{1}{c|}{0.834} &
  0.248 &
  \multicolumn{1}{c|}{\begin{tabular}[c]{@{}c@{}}0.900\\ (0.843 / 0.957)\end{tabular}} &
  \multicolumn{1}{c|}{\begin{tabular}[c]{@{}c@{}}0.888\\ (0.945 / 0.832)\end{tabular}} &
  \multicolumn{1}{c|}{0.812} &
  0.281 \\ \hline
{[}Segformer + 1x1 conv{]} + Area labels w/o refinement &
  14 / 61 &
  3.7M &
  \multicolumn{1}{c|}{\begin{tabular}[c]{@{}c@{}}0.941\\ (0.915 / 0.967)\end{tabular}} &
  \multicolumn{1}{c|}{\begin{tabular}[c]{@{}c@{}}0.945\\ (0.968 / 0.922)\end{tabular}} &
  \multicolumn{1}{c|}{0.887} &
  0.194 &
  \multicolumn{1}{c|}{\begin{tabular}[c]{@{}c@{}}0.955\\ (0.938 / 0.972)\end{tabular}} &
  \multicolumn{1}{c|}{\begin{tabular}[c]{@{}c@{}}0.952\\ (0.971 / 0.932)\end{tabular}} &
  \multicolumn{1}{c|}{0.912} &
  0.176 \\ \hline
{[}Segformer (scratch) + 1x1 conv{]} + Area labels &
  14 / 61 &
  3.7M &
  \multicolumn{1}{c|}{\begin{tabular}[c]{@{}c@{}}0.859\\ (0.745 / 0.974)\end{tabular}} &
  \multicolumn{1}{c|}{\begin{tabular}[c]{@{}c@{}}0.871\\ (0.975 / 0.767)\end{tabular}} &
  \multicolumn{1}{c|}{0.730} &
  0.349 &
  \multicolumn{1}{c|}{\begin{tabular}[c]{@{}c@{}}0.885\\ (0.793 / 0.977)\end{tabular}} &
  \multicolumn{1}{c|}{\begin{tabular}[c]{@{}c@{}}0.880\\ (0.982 / 0.777)\end{tabular}} &
  \multicolumn{1}{c|}{0.781} &
  0.329 \\ \hline
{[}Segformer (freeze) + 1x1 conv{]} + Area labels &
  14 / 61 &
  3.7M &
  \multicolumn{1}{c|}{\begin{tabular}[c]{@{}c@{}}0.817\\ (0.656 / 0.977)\end{tabular}} &
  \multicolumn{1}{c|}{\begin{tabular}[c]{@{}c@{}}0.809\\ (0.985 / 0.633)\end{tabular}} &
  \multicolumn{1}{c|}{0.649} &
  0.435 &
  \multicolumn{1}{c|}{\begin{tabular}[c]{@{}c@{}}0.830\\ (0.687 / 0.972)\end{tabular}} &
  \multicolumn{1}{c|}{\begin{tabular}[c]{@{}c@{}}0.802\\ (0.989 / 0.615)\end{tabular}} &
  \multicolumn{1}{c|}{0.681} &
  0.438 \\ \hline
{[}Segformer (freeze) + U-Net{]} + Area labels &
  14 / 63 &
  5.7M &
  \multicolumn{1}{c|}{\begin{tabular}[c]{@{}c@{}}0.918\\ (0.858 / 0.979)\end{tabular}} &
  \multicolumn{1}{c|}{\begin{tabular}[c]{@{}c@{}}0.933\\ (0.978 / 0.888)\end{tabular}} &
  \multicolumn{1}{c|}{0.843} &
  0.237 &
  \multicolumn{1}{c|}{\begin{tabular}[c]{@{}c@{}}0.942\\ (0.903 / 0.981)\end{tabular}} &
  \multicolumn{1}{c|}{\begin{tabular}[c]{@{}c@{}}0.942\\ (0.978 / 0.905)\end{tabular}} &
  \multicolumn{1}{c|}{0.885} &
  0.213 \\ \hline
{[}MobileSam{]} + Area labels &
  14 / 103 &
  10.1M &
  \multicolumn{1}{c|}{\begin{tabular}[c]{@{}c@{}}0.927\\ (0.876 / 0.977)\end{tabular}} &
  \multicolumn{1}{c|}{\begin{tabular}[c]{@{}c@{}}0.934\\ (0.975 / 0.893)\end{tabular}} &
  \multicolumn{1}{c|}{0.858} &
  0.222 &
  \multicolumn{1}{c|}{\begin{tabular}[c]{@{}c@{}}0.947\\ (0.914 / 0.980)\end{tabular}} &
  \multicolumn{1}{c|}{\begin{tabular}[c]{@{}c@{}}0.941\\ (0.979 / 0.904)\end{tabular}} &
  \multicolumn{1}{c|}{0.897} &
  0.198 \\ \hline
\end{tabular}}
\begin{tablenotes}[flushleft] \footnotesize
\item $^*$ The left and right values in "Computation time" are each measured on a standard desktop GPU (NVIDIA GeForce RTX 4090) and an onboard embedded GPU (NVIDIA Jetson Orin). The left and right values inside the brackets in "Precision" and "Recall" represent traversable and non-traversable regions, respectively. The final "Precision" and "Recall" values are calculated by averaging the two values.
\end{tablenotes}
\label{table:quantitative_eval}
\vspace{-0.25cm}
\end{table*}

The trained semantic traversability estimator's performance is compared to several prior methods demonstrated below:
\begin{itemize}
    \item {[}Segformer + 1x1 conv{]} + Point labels: The semantic segmentation model is fine-tuned with point labels obtained by projecting footsteps as in Wellhausen et al.\cite{wellhausen2019point}. Each footstep is converted to a circle with a radius of 30 pixels.
    \item Autoencoder: Following the work by Schmid et al.\cite{schmid2022reconstruct}, a denoising autoencoder modeled with convolution neural networks is trained to only reconstruct the labeled areas obtained from SAM. Areas with low reconstruction error (MSE loss smaller than 0.05) are then predicted as traversable.
    \item Segformer + Heuristic: Lightweight semantic segmentation model is directly used with heuristic category-wise traversability values \footnote{[Sidewalk, Path, Floor]: 1, [Grass, Sand, Hill, Dirt track, Land, Earth, Field]: 0.5, [Road]: 0.25} as in \cite{maturana2018offRoadNav, guan2022gaNav, roth2023viplanner}.
    \item Mask2former + Heuristic: Large semantic segmentation model is directly used with heuristic category-wise traversability values \footnote{[Pedestrian Area, Sidewalk, Lane Marking - Crosswalk]: 1, [Sand, Snow, Terrain]: 0.5, [Road, Lane Marking - General]: 0.25} as in \cite{maturana2018offRoadNav, guan2022gaNav, roth2023viplanner}.
\end{itemize}

We also compared the performance with variations of our method. Specifically, the model was fine-tuned with labels acquired without the refinement stage (i.e., only SAM is used for annotation), the model backbone was trained from scratch, the model backbone was frozen and only the adaptor was trained, the adaptor architecture was converted to a two-layer U-Net\cite{ronneberger2015unet}, and the model backbone was converted to MobileSAM\cite{zhang2023mobileSAM}, a compressed version of SAM through knowledge distillation. MobileSAM was chosen as an alternative model backbone over SAM due to its low computational cost.

\begin{figure}[t!]
\centering
\includegraphics[width=\linewidth]{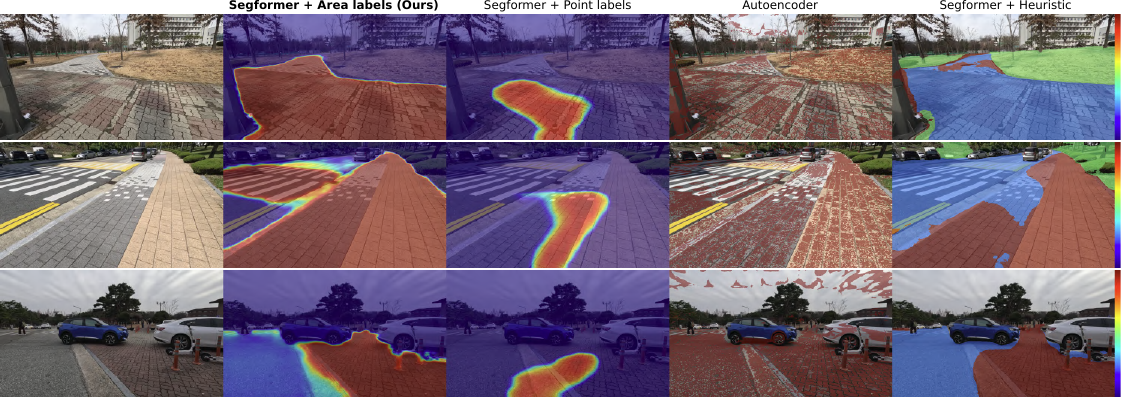}
\caption{Estimation results of semantic traversability estimators}
\vspace{-0.5cm}
\label{fig:estimation results}
\end{figure}

Using the traversability labels obtained from the proposed automatic annotation process as ground-truth labels for evaluation is inappropriate because they may contain errors. Thus, ground-truth labels were manually acquired from human annotators. 
These labels were binary (i.e., 0 or 1), based on human walking preferences.
To test the trained traversability estimator's camera viewpoint generalization capability, separate evaluations were performed for images obtained from the same and different viewpoints as the trained ones. The same viewpoint data came from a camera positioned approximately 1.36\si{m} from the ground, while the different viewpoint data was gathered between 0.4\si{m} and 0.65\si{m} from the ground (taking into account the target robot's viewpoint). A total of 1K images—500 for each viewpoint—were manually labeled. 

Table \ref{table:quantitative_eval} and Figure \ref{fig:estimation results} present the quantitative and qualitative evaluation results, respectively. Our traversability estimator, which was obtained by fine-tuning a lightweight semantic segmentation model with area labels, outperformed other methods across all metrics. It demonstrated the highest accuracy in determining traversable areas, taking into account the highest precision, recall, and IoU values, as well as the lowest RMSE. Furthermore, it demonstrated the shortest computation time, allowing for semantic traversability estimation at a high frequency of 71Hz on a desktop GPU and 16Hz on an onboard embedded GPU. The model trained with only point labels, as in \cite{wellhausen2019point}, showed low IoU values and only local area predictions due to small supervision signals during training. Concretely, the training signal is limited to dealing with extremely visually diverse image data collected in urban environments because the area indicated as traversable is significantly smaller than the area labeled as non-traversable. This demonstrates the significance of our automatic annotation method, which expands point labels to area labels through SAM, for more efficient training of image data with high visual diversity. Reconstruction-based methods \cite{schmid2022reconstruct} showed relatively low performance with large incorrect predictions due to the it's limited capability to reconstruct high-frequency information (e.g., sidewalk patterns) and discriminate traversable regions based on reconstruction errors. Directly obtaining traversability values from semantic segmentation models showed lower accuracy due to incorrect predictions for non-dominant categories in the large-scale dataset the models were originally trained on, as well as high computation time due to a large number of neural network parameters and the computationally expensive \textit{argmax} operation required for category-wise value assignments.

The traversability estimator trained using labels obtained without the refinement step performed poorly compared to those obtained with it. This is because the refinement step improves the quality of labels from SAM by converting the binary labels to fine-grained labels and removing inflated parts based on semantic information (Figure \ref{fig:stepwise annotation results}, Figure \ref{fig:qualitative_rellis_eval}). 

We also demonstrate the effectiveness of our training methodology, which involves leveraging pre-trained semantic segmentation model weights and fine-tuning them end-to-end. As shown in Table \ref{table:quantitative_eval}, models that do not use pre-trained weights or only fine-tune the added layers showed lower accuracy regardless of the added layers' size and neural network architecture. Interestingly, the model initialized with a segmentation model (i.e., MobileSAM \cite{zhang2023mobileSAM}) showed lower accuracy than ours initialized with a semantic segmentation model (i.e., Segformer \cite{xie2021segformer}). This implies that pre-trained representations of the semantic segmentation model are more suitable for semantic traversability estimation than those in the segmentation model. In our experience, representations of segmentation models (e.g., SAM, MobileSAM) are more focused on visual cues such as colors or edges rather than semantic information. Our experimental results indicate that fine-tuning the semantic segmentation model end-to-end is critical for adapting the OOD RGB features and terrain classes. Furthermore, leveraging the semantic segmentation model's pre-trained model weights allows for effective training with a small quantity of data and acts as a weak inductive bias for traversability estimation.

\begin{figure}[t!]
\centering
\scalebox{1.0}[0.9]{\includegraphics[width=\linewidth]{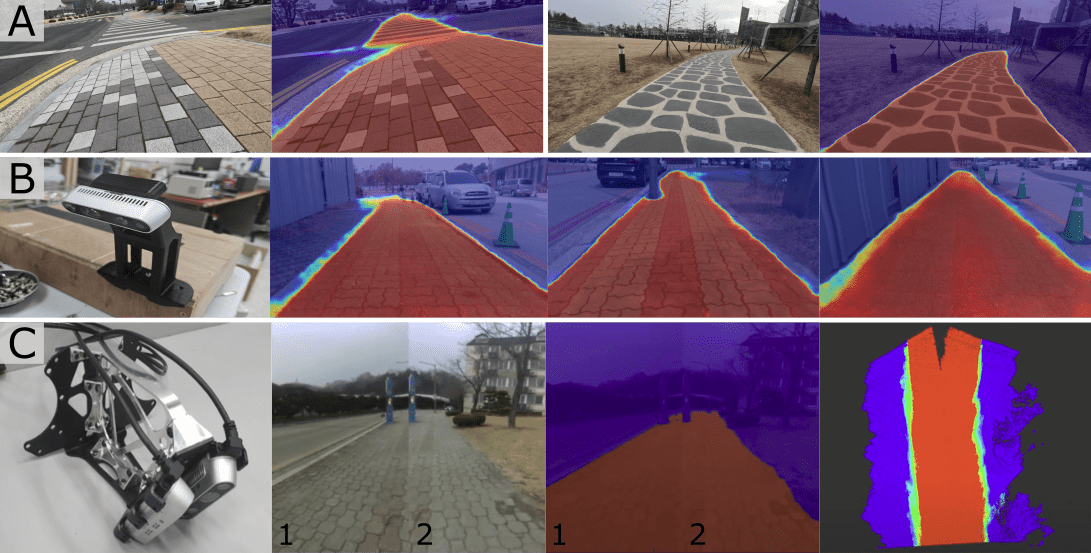}}
\caption{Estimation results for different camera viewpoints. All cameras are mounted closer to the ground than the camera mounted on a person's chest. (A: GoPro camera, B: Single RGB-D camera configured horizontally, C: Two RGB-D cameras configured vertically)}
\vspace{-0.5cm}
\label{fig:different view estimation results}
\end{figure}

The results in the "Different viewpoint" section of Table \ref{table:quantitative_eval} show that our semantic traversability estimator performed well and consistently on the test dataset containing different camera viewpoint images. This is also shown qualitatively in Figure \ref{fig:different view estimation results}-A. The trained semantic traversability estimator was further qualitatively evaluated using data from RGB-D cameras (Intel RealSense D435) with different image resolutions from the GoPro camera (Realsense: 640x360, GoPro: 1920x1080). Two camera configurations were tested: a horizontal configuration with a single camera facing forward and a vertical configuration with two cameras looking diagonally. As shown in both the image space predictions and projected 2.5D terrain maps in Figure \ref{fig:different view estimation results}, our model's predictions for all these camera configurations are highly well-aligned with humans' visual preferences. Overall, our experimental results show that our semantic traversability estimator can be generalized across different camera viewpoints, even when trained with egocentric videos, thanks to the strong data augmentation and weak inductive bias introduced from pre-trained model weights.

\begin{figure}[t!]
\centering
\scalebox{1.0}[0.9]{\includegraphics[width=\linewidth]{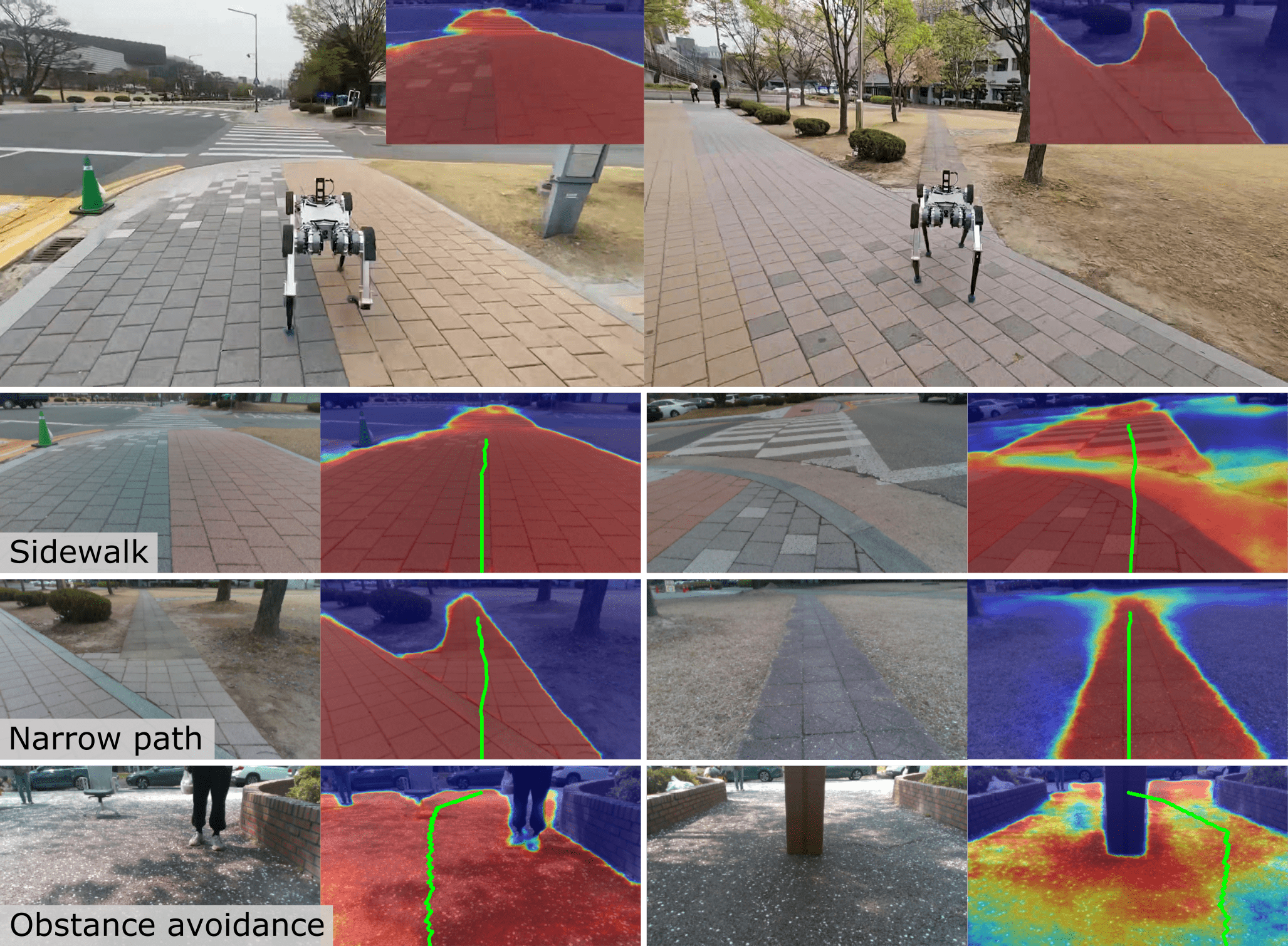}}
\caption{Path planning with traversability estimation in real-world robot deployments. The green lines are the planned paths.}
\vspace{-0.5cm}
\label{fig:real-world}
\end{figure}

\subsection{Evaluation of real-robot navigation}
We tested our semantic traversability estimator in real-robot navigation scenarios, shown in Figure \ref{fig:real-world}. The navigation experiments were conducted in several environments including sidewalks and narrow paths, where accurate traversability estimation and robust path planning are important. The quadrupedal robot \textit{Raibo2} found a path to navigate to the given goal destination considering both geometric and semantic traversability. Additionally, we performed obstacle avoidance in environments containing humans and untraversable boxes. The semantic traversability estimator identified them as untraversable objects and facilitated obstacle-overcoming path generation in conjunction with the geometric traversability.

\subsection{Evaluation of open-source dataset}
To show the applicability of our method across different domains from our custom dataset, we conducted an analysis using the Rellis3D dataset\cite{jiang2021rellis3d}. The Rellis3D dataset is an open-source dataset constructed by rolling out a wheel-based mobile robot in field environments. To apply our method to the Rellis3D setting, the time horizon $T'$ is modified from 3s to 4.5s, and segments labeled as “vegetation” are removed from the SAM mask predictions rather than “road” or “crosswalk”. These changes are made due to the difference in vehicle speed and application domain from urban to field settings. Other engineering choices and hyperparameters for both the automatic annotation and training are kept identical to those utilized for our custom dataset. For evaluation, we employed the same process and criteria as for our own dataset. Category-wise heuristic traversability values for "Segformer + Heuristic"\footnote{[Path, Grass, Sand, Dirt track, Land, Field]: 1, [Hill, Earth]: 0.5} and "Mask2former + Heuristic"\footnote{[Terrain]: 1} were updated to reflect the different environment settings.

\begin{table}[t!]
\caption{Quantitative evaluation results on the rellis3d dataset}
\centering
\resizebox{\linewidth}{!}{
\begin{tabular}{|l|c|c|c|c|}
\hline
    & Precision & Recall & IoU   & RMSE  \\ \hline\hline
\begin{tabular}[l]{@{}l@{}}\textbf{{[}Segformer + 1x1 conv{]}}\\ \textbf{+ Area labels (Ours)}\end{tabular} &
    \begin{tabular}[c]{@{}c@{}}\textbf{0.922}\\ (0.853 / 0.990)\end{tabular} &
    \begin{tabular}[c]{@{}c@{}}\textbf{0.953}\\ (0.977 / 0.929)\end{tabular} &
    \textbf{0.834} &
    \textbf{0.212} \\ \hline
\begin{tabular}[l]{@{}l@{}}{[}Segformer + 1x1 conv{]}\\ + Point labels \cite{wellhausen2019point}\end{tabular} &
    \begin{tabular}[c]{@{}c@{}}0.867\\ (0.975 / 0.758)\end{tabular} &
    \begin{tabular}[c]{@{}c@{}}0.657\\ (0.316 / 0.997)\end{tabular} &
    0.312 &
    0.460 \\ \hline
Segformer + Heuristic \cite{maturana2018offRoadNav, guan2022gaNav, roth2023viplanner} &
    \begin{tabular}[c]{@{}c@{}}0.819\\ (0.652 / 0.985)\end{tabular} &
    \begin{tabular}[c]{@{}c@{}}0.870\\ (0.967 / 0.773)\end{tabular} &
    0.640 &
    0.395 \\ \hline
Mask2former + Heuristic \cite{maturana2018offRoadNav, guan2022gaNav, roth2023viplanner} &
    \begin{tabular}[c]{@{}c@{}}0.919\\ (0.868 / 0.971)\end{tabular} &
    \begin{tabular}[c]{@{}c@{}}0.935\\ (0.930 / 0.939)\end{tabular} &
    0.807 &
    0.232 \\ \hline
\begin{tabular}[l]{@{}l@{}}{[}Segformer + 1x1 conv{]}\\ + Area labels w/o refinement\end{tabular} &
    \begin{tabular}[c]{@{}c@{}}0.871\\ (0.748 / 0.994)\end{tabular} &
    \begin{tabular}[c]{@{}c@{}}0.912\\ (0.986 / 0.837)\end{tabular} &
    0.737 &
    0.308 \\ \hline
\end{tabular}}
\label{table:quantitative_rellis_eval}
\end{table}

\begin{figure}[t!]
\centering
\includegraphics[width=\linewidth]{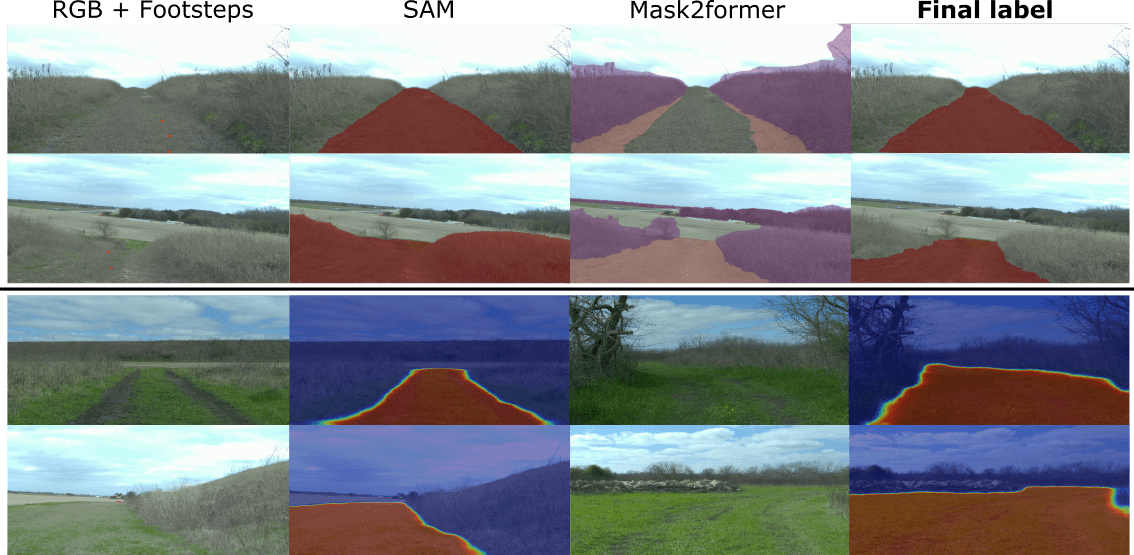}
\caption{Qualitative results on Rellis3D dataset. The top two lines display the stepwise annotation results, while the bottom two lines show the estimated outputs from the trained traversability estimator. In Mask2former predictions, only terrains and vegetations are color-segmented with 
brown and purple if detected.}
\vspace{-0.35cm}
\label{fig:qualitative_rellis_eval}
\end{figure}

The quantitative and qualitative evaluation results on the Rellis3D dataset are shown in Table \ref{table:quantitative_rellis_eval} and Figure \ref{fig:qualitative_rellis_eval}. Our traversability estimator performed better than other approaches on all metrics, which is consistent with the experimental results exhibited in our custom dataset. Furthermore, our two-step automatic annotation pipeline enables to generate clear traversability labels on areas where either SAM or Mask2former alone cannot accurately predict. Interestingly, the introduction of the semantic refinement stage in the annotation process significantly improved the traversability estimator's performance when compared to urban situations. This is because, in field situations, SAM frequently generates inflated labels due to ambiguous visual cues. The filtering process based on the predictions from the large semantic segmentation models enables to alleviate the overestimation problem of SAM and produce more reliable labels.

Our experimental results and analysis on the Rellis3D dataset show that our methodology is not limited to urban settings or our custom dataset, but can also be easily extended to other environment settings (e.g., field, park) or open-source datasets, allowing for the training of highly accurate and computationally light semantic traversability estimators.

\section{Conclusion}

\begin{figure}[t!]
\centering
\includegraphics[width=\linewidth]{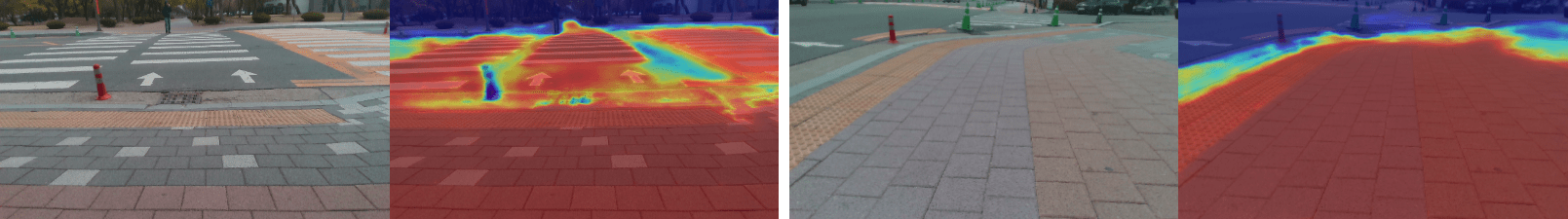}
\caption{Limitations of the proposed semantic traversability estimator. (Left) Occasionally, speed bumps are misidentified as crosswalks, leading to incorrect predictions of traversability. (Right) Crosswalks located at a distance are sometimes not recognized as traversable.}
\vspace{-0.5cm}
\label{fig:limitation}
\end{figure}

We have proposed an effective methodology for training semantic traversability estimator with a dataset constructed with egocentric videos and an automated annotation strategy. We show that leveraging egocentric videos for robot navigation allows for scalable and efficient data acquisition. Extensive quantitative and qualitative experiments demonstrate that the yielded traversability estimator is highly accurate, camera viewpoint and configuration generalizable, and lightweight. Further deployment for autonomous quadruped robot navigation in an urban environment showcases the method’s capability for real-world applications.

As shown in Figure \ref{fig:limitation}, the proposed semantic traversability estimator sometimes encounters limitations when visual features are not clearly identifiable, such as with similarly appearing objects or distant objects. However, these limitations do not significantly impact navigation scenarios, as the visual features become clearer as the robot approaches.

Promising future directions include training a global semantic traversability estimator by scaling the dataset (e.g., worldwide data collection, leveraging internet-scale dataset\cite{grauman2022ego4d}) and improving the data noise handling by incorporating representation-based methods\cite{frey2023fastTrav, seo2023pseudoLabel}.

\bibliographystyle{IEEEtran}
\bibliography{IEEEabrv, ref}
\end{document}